\documentclass[11pt,letterpaper]{article}
\usepackage{coling2016}
\usepackage{times}
\usepackage{url}
\usepackage{latexsym}
\usepackage{graphicx}
\usepackage{multirow}
\usepackage{amsmath}
\usepackage{amsfonts}
\usepackage{booktabs}
\usepackage{algorithm}
\usepackage{algpseudocode}

\title{Distance Metric Learning for Aspect Phrase Grouping}

\author{Shufeng Xiong$^{1,3}$, Yue Zhang$^2$, Donghong Ji$^1$\thanks{~~corresponding author}, Yinxia Lou$^1$ \\
  $^1$Computer School, Wuhan University, Wuhan, China \\
  $^2$Singapore University of Technology and Design, Singapore \\
  $^3$Pingdingshan University, Pingdingshan, China \\
  {\tt \{xsf, dhji\}@whu.edu.cn} \\
  {\tt yue\_zhang@sutd.edu.sg} \\}

\date{}

\begin{document}
\maketitle
\begin{abstract}
Aspect phrase grouping is an important task in aspect-level sentiment analysis. It is a challenging problem due to polysemy and context dependency. We propose an Attention-based Deep Distance Metric Learning (ADDML) method, by considering aspect phrase representation as well as context representation. First, leveraging the characteristics of the review text, we automatically generate aspect phrase sample pairs for distant supervision. Second, we feed word embeddings of aspect phrases and their contexts into an attention-based neural network to learn feature representation of contexts. Both aspect phrase embedding and context embedding are used to learn a deep feature subspace for measure the distances between aspect phrases for K-means clustering. Experiments on four review datasets show that the proposed method outperforms state-of-the-art strong baseline methods.
\end{abstract}

\section{Introduction}
\label{intro}
\blfootnote{
     \hspace{-0.65cm}  
     This work is licenced under a Creative Commons 
     Attribution 4.0 International License.
     License details:
     \url{http://creativecommons.org/licenses/by/4.0/}
}

For aspect-level sentiment analysis \cite{HuLiu-1446,PangLee-1450}, aspect identification from the corpus is a necessary step.
Here {\em aspect} is the name of a feature of the product, while an {\em aspect phrase} is a word or phrase that actually appears in a sentence to indicate the aspect.
Different aspect phrases can be used to describe the same aspect.
For example, ``picture quality'' could be referred to ``photo'', ``image'' and ``picture''.
All aspect phrases in the same group indicate the same aspect.
In this paper, we assume that all aspect phrases have been identified by using existing methods \cite{JinHo-1452,KobayashiInui-1453,KimHovy-1125}, and focus on grouping domain synonymous aspect phrases.

Most existing work employed unsupervised methods, exploiting lexical similarity from semantic dictionary as well as context environments \cite{ZhaoHuang-1265,ZhaiLiu-1086,GuoZhu-1250}.
The context for an aspect phrase is formed by aggregating related sentences that mention the same aspect phrase.
Thereafter, aspect phrase and context environment are represented using bag-of-word (BoW) models separately, and integrated into a unified learning framework.

One limitation of the existing methods is that they do not model the interaction between aspect phrases and their contexts explicitly. For example, in the review ``the picture is clear, bright and sharp and the sound is good'', the words ``clear'', ``bright'' and ``sharp'' are related to the aspect phrase ``picture'', while the word ``good'' is related to the aspect phrase ``sound''. By the traditional model, these words are not differentiated when they are taken for the context, thereby causing noise in the grouping of the two aspect phrases.

To address this issue, we propose a novel neural network structure that automatically learns the relative importance of each context word with respect to a target/aspect phrase, by leveraging an attention model \cite{LuongPham-1792,RushChopra-1794,LingTsvetkov-1793}. As shown in Figure 1, given a sentence that contains an aspect phrase, we use a neural network to find a vector representation of the aspect phrase and its context. For the grouping of a certain aspect phrase, we concatenate all the occurrences of the aspect phrase in a corpus to find its vector form. Thus, the problem of aspect phrase grouping is transformed into a clustering problem in the resulting vector space. Different from traditional methods, which leverage a bag-of-word feature space, our vector space considers not only words, but also semantic similarities between aspect phrases and contexts \cite{XuWang-1779}.

\begin{figure}[!tb]
\center
\includegraphics[width=0.49\textwidth]{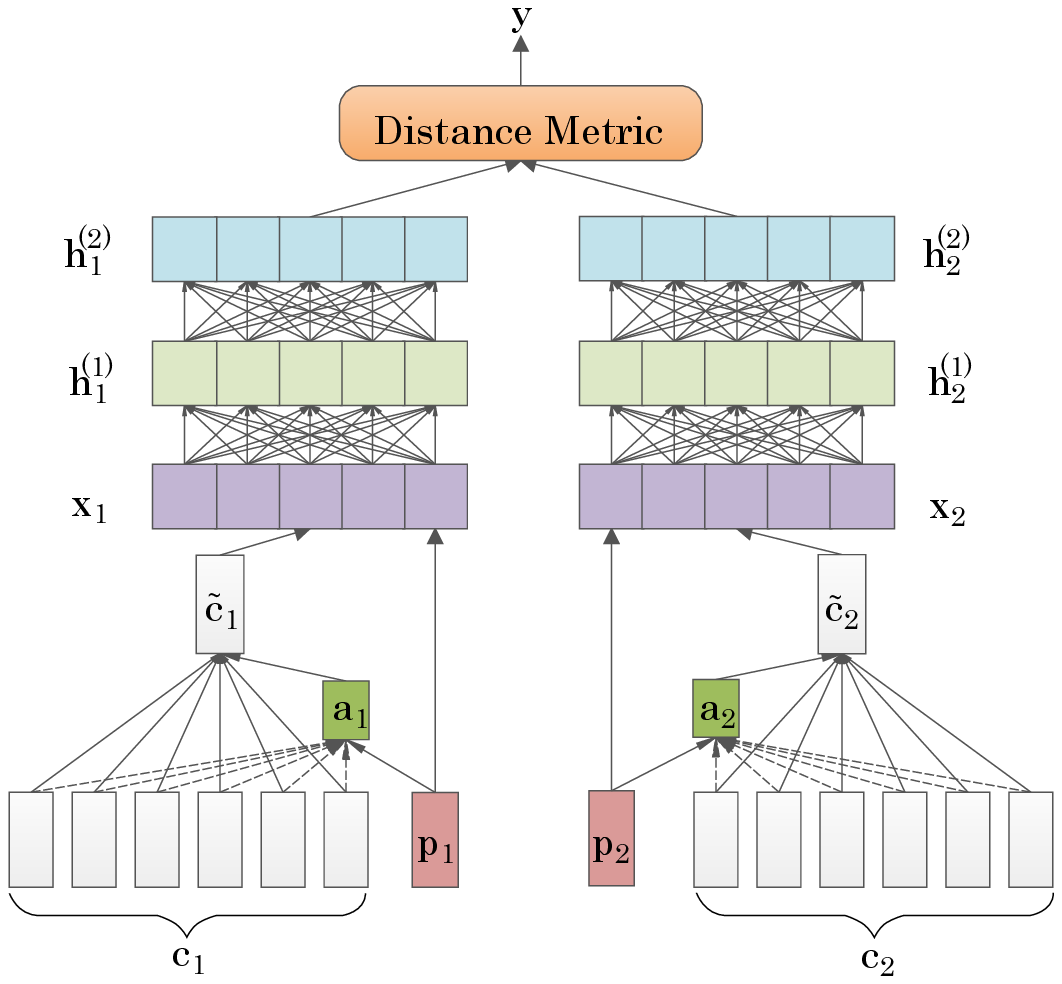}
\caption{Architecture of the proposed method. For a given pair of aspect phrases $p_1$ and $p_2$, with their contexts $c_1$ and $c_2$ respectively, two vectors $x_1$ and $x_2$ are obtained via attention-based semantic combination, and then mapped into the same feature subspace as $h_1^{(2)}$ and $h_2^{(2)}$.}
\label{fig:example}
\end{figure}

One challenge to the success of our method is the finding of a proper training algorithm for the neural network model. Inspired by word embedding training methods \cite{CollobertWeston-1507,MikolovSutskever-1335}, we take a negative sampling approach. In particular, we take pairs of sentences that contain the same aspect phrase as positive training examples, and pairs of sentences that contain incompatible aspect phrase as negative training examples, maximizing a score margin between positive and negative examples. Here two aspect phrases are incompatible if the distance based on a semantic lexicon is large \cite{FaruquiDodge-1548,YuDredze-2011}.

To find a better vector space representation, we add two nonlinear transformation layers, as shown in  $h^{(1)}$ and  $h^{(2)}$ in Figure 1. This method is similar to the Mahalanobis distance metric learning for face verification \cite{HuLu-1683}. Model training is performed by back-propagation over all neural nodes. With such vector space being learned, direct K-means clustering can be used to group aspect phrases.

Results on a standard benchmark show that our neural network significantly outperforms traditional models. The average results on 4 domains reached 0.51 (Purity) and 1.74 (Entropy), better than the previous best result (0.43 in Purity and 2.02 in Entropy).

\section{Method}
Our proposed model addresses two main problems: (1) to express fine-grained semantic information with a fixed length vector, which can naturally combine aspect phrases and their contexts, and (2) to provide nonlinear transformations to learn a feature subspace, under which the distance between each intra-group aspect phrase pair is smaller than that between each inter-group pair.
We discuss an attention-based semantic composition model in Section 2.1, and then describe a Multi-Layer Perceptron (MLP)-based nonlinear transformation model in Section 2.2, which are designed for (1) and (2), respectively.

\subsection{Attention-Based Semantic Composition}

The goal of the model is to learn the semantic representation of the context of each aspect phrase.
For our task, the same context is frequently shared by more than one aspect phrases. For example, in the sentence:  ``the \textbf{picture} is clear and sharp and the \textbf{sound} is good.", two different aspect phrases ``picture" and ``sound" are mentioned. We use an attention-based neural semantic composition model to consider contextual words based on their weight scores with respect to each aspect phrase. In particular, given each word vector $e_i$, which is projected into a word embedding matrix $L_w \in \mathbb{R}^{d \times \left | V \right |}$, where $d$ is the dimension of word vector and $\left | V \right |$ is the size of word vocabulary. All of $e_i$ can be randomly initialized from a uniform distribution, and then updated during the back propagation training procedure.
Alternatively, another way is using pre-trained vectors as initialization, which is learnt from text corpus with embedding learning algorithms.
In our experiment, we adopt the latter strategy.
Let $c=\{e_i|e_i \in R^{l \times 1}\}_{i=1,2,...,n}$ denote the set of $n$ input words in context, where $l$ is the dimension of the original context segment.
We employ a linear layer to combine the original context vector $c$ and attentional weight $a$ to produce an attentional context representation as:
\begin{equation}
\tilde{c} = f_w(c,a) ,
\end{equation}
where $f_w$ is a weighted average function.
The idea is to give different weights for different words in the context when deriving the context vector $\tilde{c}$.
The weight $a \in \mathbb{R}^{n \times 1}$ is a variable-length attention vector, whose size is equal to the number of words in the context.
Its value is computed as follows:
\begin{equation}
a(e_i) = \frac{\exp (score(e_i,p))}{\sum_{i'} {\exp (score(e_{i'},p))}} ,
\end{equation}
where $score(e_i,p)=W_a^T[e_i;p]$ and $W_a \in \mathbb{R}^{(2*d) \times 1}$ is a model parameter to learn.
Although the length of context is variable, our model uses a fixed-length $W_a$ parameter to weight the importance of each word $e_i$ for its corresponding aspect phrase $p$. This results in a fixed length vector form for each aspect phrase in a variable-size context.

\begin{figure*}[!tb]
\center
\includegraphics[width=\textwidth]{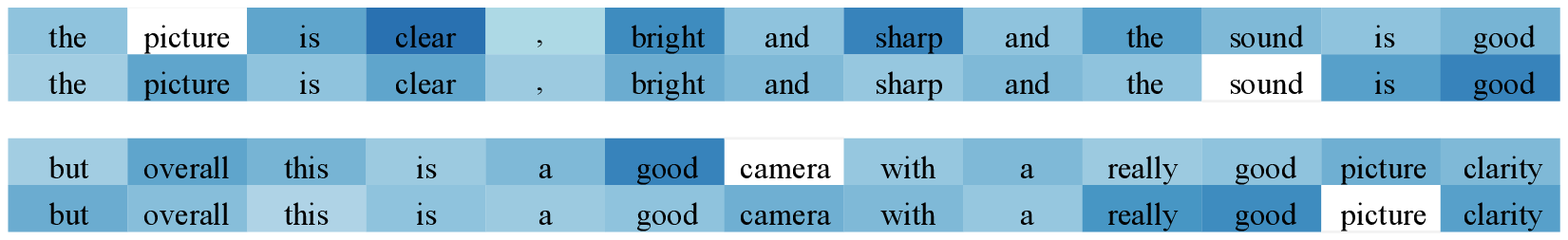}
\caption{Visualization of weighted context by our attention model. There is a context which contains two aspect phrases (white background), where the weight of each word is taken from the developing dataset. For different aspect phrases, the words in the same context have different weights. Different background represent different weights, bigger weights corresponding to deeper background. }
\label{fig:example}
\end{figure*}

\subsection{MLP-Based Nonlinear Transformation}
After obtaining the attention-based context $\tilde{c}$, we employ a MLP-based nonlinear transformation to learn a feature subspace for final aspect phrase grouping.
Although $\tilde{c}$ is a weighted context according to the aspect phrase, the aspect phrase $p$ itself is still a necessary source of information for grouping.
Therefore, we concatenate $\tilde{c}$ and $p$ to produce a vector $x$ as the input to MLP.

Our model is based on a variant of the Mahalanobis distance metric learning method \cite{HuLu-1683}.
The problem is formulated as follows.
Given a training set $X=\{x_i|x_i \in R^{(2*d) \times 1}\}_{i=1,2,...,m}$, where $x_i$ is the $i$th training sample and $m$ is the size of training set.
The method aims to seek a linear transformation $W$, under which the distance between any two samples $x_i$ and $x_j$ can be computed as:
\begin{equation}
d_w(x_i,x_j) = \left \| Wx_i-Wx_j \right \|_2
\label{dis}
\end{equation}
where $W$ is an alternative of the covariance matrix $M$ in Mahalanobis distance. $\|A\|_2$ represents the $L2$ norm of the matrix $A$.
$M$ can be decomposed by:
\begin{equation}
M=W^TW,
\end{equation}
Further, Equation (3) can be rewritten as
\begin{equation}
\label{maha}
\begin{split}
d_w(x_i,x_j) & = \left \| Wx_i-Wx_j \right \|_2 \\
& = \sqrt{(x_i-x_j)^TW^TW(x_i-x_j)} \\
& = \sqrt{(x_i-x_j)^TM(x_i-x_j)}
\end{split}
\end{equation}
Equation (\ref{maha}) is the typical form of Mahalanobis distance between $x_i$ and $x_j$.
Therefore, Equation (\ref{dis}) is both the Euclidean distance of two samples in the linear transformed space, and the Mahalanobis distance in the original space.
The transformation $Wx$ can be replaced with a generalized function $g$.
When $g$ is a nonlinear function, we obtain the nonlinear transformation form of Mahalanobis distance.
Following \newcite{HuLu-1683}, we use the squared Euclidean distance in our model:
\begin{equation}
d_g^2(x_i,x_j) = \left \|g(x_i)-g(x_j) \right \|_2^2
\label{euc}
\end{equation}

As shown in Figure 1, we use hierarchical nonlinear mappings to project the samples to a feature subspace.
Assume that there are $M$ layers in the designed network, and $k^{(m)}$ units in the $m$th layer, where $m=1,2,...,M$.
For a given aspect phrase sample $x$, the output of the first layer is computed as
\begin{equation}
h^{(1)} = f_a(W^{(1)}x + b^{(1)}) \in \mathbb{R}^{k^{(2)}},
\end{equation}
where the weight matrix $W^{(1)} \in \mathbb{R}^{k^{(2)} \times k^{(1)}}$ can be seen as a linear projection transformation, $b^{(1)} \in \mathbb{R}^{k^{(2)}}$ is a bias vector, and $f_a: \mathbb{R} \mapsto \mathbb{R}$ is a nonlinear activation function.

Subsequently, the output of the first layer $h^{(1)}$ is used as the input of the second layer.
In the same way, the output of the second layer is
\begin{equation}
 h^{(2)} = f_a(W^{(2)}h^{(1)} + b^{(2)}) \in \mathbb{R}^{k^{(3)}},
\end{equation}
where $W^{(2)} \in \mathbb{R}^{k^{(3)} \times k^{(2)}}$, $b^{(2)} \in \mathbb{R}^{k^{(3)}}$ and $f_a$ are the projection matrix, a bias and a nonlinear activation function of the second layer, respectively.

Finally, the output of the topmost layer is calculated as follows:
\begin{equation}
h^{(M)} = f_a(W^{(M)}h^{(M-1)} + b^{(M)}) \in \mathbb{R}^L,
\end{equation} 
where $L$ is the dimension of the output vector.

Given a pair of aspect phrase samples $x_i$ and $x_j$, let $g(x_i) = h_i^{(M)}$ and $g(x_j) = h_j^{(M)}$.
The function $g$ represents a hierarchical nonlinear transformation, in which the aspect phrase sample pair is passed through the $M$-layer deep network and mapped into a feature subspace.
By using Equation (\ref{euc}), we can measure the distance between the sample pair in the new feature subspace.

\subsection{K-means Clustering}
With a given corpus, we first employ the parallel deep neural network to learn the semantic representation $h^{(M)}$, and then utilize the K-means algorithm to perform clustering of $h^{(M)}$. During training, we sample aspect phrase pairs using the sentence as context. During testing, we concatenate all the sentences that mention the aspect phrase for clustering the aspect phrase.

\subsection{Model Training}

The ultimate goal of our model is to make the distance metric effective for grouping aspect phrase samples.
To achieve this, we use a large-margin framework to restrict the distance, as proposed by \newcite{MignonJurie-1865}.
In particular, sample pairs containing the same aspect phrase are used as positive instances and sample pairs with incompatible aspect phrase are used as negative instances.

We exploit lexical similarity to obtain incompatible aspect phrases, which have low similarity in semantic lexicon.
In particular, we choose WordNet as the semantic lexicon.
Two aspect phrase are incompatible when the WordNet similarity between them is smaller than a threshold $\eta$.
And the WordNet similarity is calculated by Equation (\ref{eq:jcn}) 
\begin{equation}
Res(w_1,w_2)=IC(LCS(w_1,w_2))
\end{equation}
\begin{equation}
IC(w)=-logP(w)
\end{equation}
\begin{equation}
Jcn(w_1,w_2)=\frac{1}{IC(w_1)+IC(w_2)-2 \times Res(w_1,w_2)}
\label{eq:jcn}
\end{equation}
where LCS (lease common subsumer) is the most specific concept that is a shared ancestor of the two concepts represented by the words \cite{Pedersen-1471}. $P(w)$ is the probability of the concept word $w$. In our experiments, the threshold is set to 0.85.

Traditional methods \cite{ZhaiLiu-1272,ZhaiLiu-1459} exploit lexical knowledge to provide soft constraint for clustering aspect phrases. They assume that the aspect phrases that have high similarity in semantic lexicon, are likely to belong to the same group.
In this cause, our method uses a similar assumption. 

For obtaining the training data, we apply an extra sample pair generation process.
The generated sample aspect phrase pairs are fed into left and right sub neural network of Figure 1, respectively.
Specifically, each training sentence is utilized with its labelled aspect phrase as a gold sample.
Then, we combine each aspect phrase and its related sentences to form training sample set.
For example, given an aspect phrase $p_1$ and a sentence set $S^1 = \{s^1_1, s^1_2,...,s^1_m\}$, in which there are $m$ sentences that mention $p_1$, we can construct $m$ samples $\{p_1 \cup s^1_1,p_1 \cup s^1_2,...,p_1 \cup s^1_m\}$.
The group label of the sample is the same as its original aspect phrase, e.g. when $p_1$ belongs to group $1$, then all of $p_1 \cup s^1_i$ have the group label $1$. 

Assuming that there are $n$ selected aspect phrases, the number of positive sample pairs candidates is $\binom{n}{2}$. A negative sample pair is formed by randomly selecting incompatible aspect phrases and their contexts.
For balancing the training set, we obtain the same number of negative sample pairs.

In the training objective, the distance $d_g^2(x_i,x_j)$ of positive instances ($l_{ij} = 1$) is less than a small threshold $t_1$ and that of negative instance ($l_{ij} = -1$) is higher than a large threshold $t_2$, where the label $l_{ij}$ denotes the similarity or dissimilarity between a sample pair $x_i$ and $x_j$, and $t_2 > t_1$. Let  $t > 1$, $t_1 = t - 1$ and  $t_2 = t + 1$.
This constraint can be formulated as follows:
\begin{equation}
l_{ij}(t - d_g^2(x_i,x_j)) > 1,
\label{cons}
\end{equation}
Equation  (\ref{cons}) enforces the margin between $d_g^2(x_i,x_j)$.

During the training phase, each aspect phrase pair must satisfy the constraint in Equation  (\ref{cons}).
Let $\omega = 1 - l_{ij}(t - d_g^2(x_i,x_j))$, we minimize the objective function:
\begin{equation}
J = \frac{1}{2}\sum_{i,j} \sigma (\omega) + \frac{\lambda}{2} \sum_{m=1}^M (\left \|W^{(m)} \right \|_F^2 + \left \|b^{(m)} \right \|_2^2)
\label{obj}
\end{equation}
where $\sigma(\omega) = \frac{1}{\beta} \log (1 + \exp (\beta \omega))$ is the generalized logistic loss function \cite{MignonJurie-1865}, which is a smooth approximation of the hinge loss $E(z) = \max (0,z)$. $\beta$ is the sharpness parameter, $\lambda$ is a regularization parameter and $\left \|W\right \|_F^2$ represents the Frobenius norm of matrix $W$.

The minimization problem in Equation (\ref{obj}) is solved using a stochastic sub-gradient descent scheme.
We train the network using back-propagation.
We set the dimension of word vectors as 200, the output length of MLP as 50.
The parameters of the linear layer are initialized using normalized initialization \cite{Glorot-1811}.
We train a three-layer MLP and employ dropout with 50\% rate to the hidden layer.
We choose $tanh$ as the activation function.
The threshold $t$, the regularization parameter $\lambda$ and learning rate $\mu$ are empirical set as 3, 0.002 and 0.03 for all experiments, respectively.

\section{Experiments}
\label{exp}

\subsection{Data Preparation}

We employ the datasets of \newcite{XiongJi-1867} to evaluate our proposed approach. The datasets are based on \textit{Customer Review Datasets} (CRD) \cite{HuLiu-1446}, including four different domains: digital camera (DC), DVD player, MP3 player (MP3) and cell phone (PHONE). We take 3 different random splits of datasets (30\% train, 50\% test, and 20\% development). The statistics are described in Table \ref{tab:ds1}.

\begin{table}[!htb]
\center
\begin{tabular}{lcccc}
\toprule 
  & DC & DVD & MP3 & PHONE\\ 
\midrule 
\#Sentences & 330 & 247 & 581 & 231 \\ 
\#Aspect Phrases & 141 & 109 & 183 & 102 \\ 
\#Aspects & 14 & 10 & 10 & 12\\  
\#Pairs & 19163 & 11211 & 64945 & 8855\\ 
\bottomrule
\end{tabular}
\caption{Statistics of the review corpus. \# donotes the size\label{tab:ds1}}
\end{table}

\subsection{Pre-trained Word Vectors}
We use \textit{Glove}\footnote{http://nlp.stanford.edu/projects/glove/} tools to train word embeddings, and the training parameters are set by following \newcite{PenningtonSocher-1333}. Because of the review corpus is too small for learning word embeddings, we use \textit{Amazon Product Review Data} \cite{JindalLiu-1866} as one auxiliary training corpus.

\subsection{Evaluation Metrics}
Since the problem of aspect phrase grouping is a clustering task, two common measures for clustering are used to its evaluate performance \cite{ZhaiLiu-1272} : \textit{Purity} and \textit{Entropy}.


\textbf{Purity} is the percentage of correct clusters that contain only data from the gold-standard partition. A large Purity reflects a better model.

\textbf{Entropy} looks at how various groups of data are distributed within each cluster. A smaller Entropy reflects a better model. 
%
%

\subsection{Baseline Methods and Settings}
The proposed ADDML method is compared with a number of methods, which include (1) the state-of-the-art methods, (2) baseline neural methods.

In the first category\footnote{Because we use sample pairs but not cluster label for training, it is not possible to train a supervised classifier for testing.}, all methods except Kmeans and CC-Kmeans exploit labelled data, which is generated using sharing word constraint and lexical similarities based on WordNet\footnote{The generation of labelled data follows \newcite{ZhaiLiu-1086}.}:

\textbf{Kmeans}. The most popular clustering algorithm based on distributional similarity with \textit{cosine} as similarity measure and BoW as feature representation.

\textbf{DF-LDA}. A combination of Dirichlet Forest Prior and LDA model, in which it can encode domain knowledge (the label) into the prior on topic-word multinomials \cite{AndrzejewskiZhu-1606}\footnote{There are other LDA based methods for this task, such Constrained LDA \cite{ZhaiLiu-1272}. Although Constrained LDA used two domains data form CRD dataset as core corpus, they actually crawled many other camera and phone reviews. So we are unable to compare with them by directly using their published results. Since DF-LDA is more effectiveness and suitable for our smaller datasets, we use DF-LDA as the representative of the LDA-based methods.}. The code is available in author's website\footnote{http://pages.cs.wisc.edu/\~{}andrzeje/research/df\_lda.html}.

\textbf{L-EM}. A state-of-the-art semi-supervised method for clustering aspect phrases \cite{ZhaiLiu-1086}. It employed lexical knowledge to provide a initialization for EM. We implemented this method ourselves.

\textbf{CC-Kmeans}. It is proposed by \newcite{XiongJi-1867}, it encodes the capacity limitation as constraint and proposes a capacity constrained K-means to cluster aspect phrases. We use the code from the author\footnote{https://github.com/pdsujnow/cc-kmeans}.

The word embedding composite methods employ different composite strategies to form the sample vector, respectively. The clustering method is Kmeans with cosine distance in which word embedding is used as feature vector.

\textbf{AVG/MIN/MAX+MLP} use the average/minimum/maximum value of all the context word vectors in each dimension as the context vector $\tilde{c}$, respectively, and then, concatenates aspect phrase $p$ and $\tilde{c}$ to form the sample vector.

\textbf{AP} only uses aspect phrase (AP) vector to cluster aspect phrases.

Since all the methods based on Kmeans depend on the random initiation, we use the average results of 10 runs as the final result. For L-EM, we use the same parameter settings with the original paper.

\subsection{Results}

\begin{table*}[!htb]
\center
\resizebox{\textwidth}{!}{
\begin{tabular}{lcccccccccc}
\toprule 
\multirow{2}{*}{} &
\multicolumn{5}{c}{Purity} & 
\multicolumn{5}{c}{Entropy}\\ 
\cmidrule(lr){2-6} \cmidrule(lr){7-11}
& DC & DVD & MP3 & PHONE & \textbf{avg} & DC & DVD & MP3 & PHONE & \textbf{avg}\\
\midrule
Kmeans & 0.4079 & 0.3922 & 0.3509 & 0.3333 & 0.3711 & 2.2627 & 2.0056 & 2.2862 & 2.5894 & 2.2860 \\ 
DF-LDA     & 0.4365 & 0.4362 & 0.3467 & 0.4329 & 0.4132 & 2.1355 & 1.9705 & 2.2054 & 2.3875 & 2.1747 \\ 
L-EM       & 0.4605 & 0.4706 & 0.3333 & 0.4561 & 0.4301 & 2.0451 & 1.9145 & 2.2427 & 1.8952 & 2.0244 \\ 
CC-Kmeans  & 0.4554 & 0.4483 & 0.3333 & 0.4353 & 0.4181 & 1.9604 & 1.9841 & 2.2897 & 1.8794 & 2.0284 \\ 
\midrule
AVG        & 0.5089 & 0.4483 & 0.3667 & 0.4941 & 0.4545 & 1.7203 & 2.1759 & 2.2030 & 1.7039 & 1.9508 \\ 
MIN        & 0.4554 & 0.3218 & 0.3600 & 0.4118 & 0.3872 & 2.1055 & 2.5479 & 2.6598 & 2.2158 & 2.3822 \\ 
MAX        & 0.4554 & 0.3563 & 0.3667 & 0.4353 & 0.4034 & 2.1230 & 2.4440 & 2.6036 & 2.1744 & 2.3363 \\ 
AP         & 0.4196 & 0.4253 & 0.3600 & 0.4588 & 0.4159 & 2.1816 & 2.2074 & 2.3087 & 1.9946 & 2.1731 \\
\midrule
ADDML      & \textbf{0.5658} & \textbf{0.5098} & \textbf{0.3684} & \textbf{0.6143} & \textbf{0.5146} & \textbf{1.7119} & \textbf{1.8043} & \textbf{2.1274} & \textbf{1.3282} & \textbf{1.7429} \\ 
\bottomrule
\end{tabular}}
\caption{Comparison of Purity and Entropy with baselines. Our model is ADDML.
\label{tab:rs1}}
\end{table*}


We present and compare the results of ADDML and the 8 baseline methods based on 4 domains. The results are shown in Table 2, where \textbf{avg} represents the average result of the 4 domains. The results are separated into two groups according to categories of the baseline methods. Our approach outperforms baseline methods on the average result of all domains. In addition, we make the following observations:
\begin{itemize}
\item From the first group, we can see that L-EM and CC-Kmeans perform better than the other methods. The methods that exploit external knowledge and constraint can achieve better performance. However, the proposed ADDML method outperforms all baselines by using weighted contexts as well as distance metric learning.

\item From the second group, all methods employ word embeddings to represent word semantic and text composition semantic. Yet these methods achieve uneven results due to different semantic composition strategies. The neural bag-of-word AVG method performs better than the others in the overall result, in which it averages the semantics of each word in the context. The average operation is a commonly used approach in many neural methods, such as CNN (Convolution Neural Network), and achieves better performance. However, it still falls behind our ADDML method according to its task-independent characteristic.

\end{itemize}

\subsection{Discussion}


\textbf{Case study} ~We manually examined a number of samples, which can be successfully grouped by ADDML but not the baselines. For example, two aspect phrases "\textit{photo quality}" and "\textit{quality}" belong to group "\textit{picture}" and "\textit{build quality}", respectively. Most of the baselines incorrectly clustered them into the same group, while ADDML correctly grouped them. There are two main reasons: (1) the two aspect phrase themselves have similar semantics characteristic and share words, (2) reviewers commonly used similar words to express their opinion. ADDML can learn an exact vector representation that are context sensitive, while the baseline methods can not distinguish similar contexts. Figure 2 shows some example results of attention values.

\textbf{Module Analysis}  ~ADDML has three modules: attention-based semantic composition module ($atn$), MLP-based nonlinear transformation module ($mlp$) and metric learning ($ml$). For studying the contribution of each module, we introduce a general convolution neural network ($cnn$) as an alternative to $atn$. $cnn$ is a state-of-the-art neural network method for modelling semantic representation of sentence \cite{KalchbrennerGrefenstette-1385,TangQin-1755}, which extracts N-gram features by convolution operations.

\begin{table*}[!tb]
\resizebox{\textwidth}{!}{
\begin{tabular}{lcccccccccccc}
\toprule 
\multirow{2}{*}{} &
\multicolumn{6}{c}{Purity} & 
\multicolumn{6}{c}{Entropy}\\ 
\cmidrule(lr){2-7} \cmidrule(lr){8-13}
& DC & DVD & MP3 & PHONE & \textbf{avg} &  $\uparrow$ & DC & DVD & MP3 & PHONE & \textbf{avg} & $\uparrow$ \\
\midrule
AP         & 0.4196 & 0.4253 & 0.3600 & 0.4588 & 0.4159 & & 2.1816 & 2.2074 & 2.3087 & 1.9946 & 2.1731 & \\
\midrule
$cnn+ml$      & 0.4673 & 0.4609 & 0.3003 & 0.4947 & 0.4308 & 3.6\% & 1.8984 & 1.9515 & 2.2908 & 1.6294 & 1.9433 & 10.6\% \\ 
$atn+ml$       & 0.4605 & 0.4706 & 0.3070 & 0.5088 & 0.4367 & 5.0\% & 1.8477 & 1.8158 & 2.2662 & 1.5179 & 1.8619 & 14.3\% \\ 
$cnn+mlp+ml$    & 0.5526 & 0.4118 & 0.3246 & 0.6140 & 0.4757 & 14.4\% & 1.8494 & 1.8980 & 2.1626 & 1.3336 & 1.8109 & 16.7\% \\ 
$atn+mlp+ml$(ADDML)	& \textbf{0.5658} & \textbf{0.5098} & \textbf{0.3684} & \textbf{0.6143} & \textbf{0.5146} & 23.7\% & \textbf{1.7119} & \textbf{1.8043} & \textbf{2.1274} & \textbf{1.3282} & \textbf{1.7429} & 19.8\% \\ 
\bottomrule
\end{tabular}}
\caption{The result of different module combinations.
\label{tab:rs2}}
\end{table*}

Table 3 reports the results of different module combinations. We use AP, which only uses aspect phrase vector for clustering, as a reference. The symbol  $\uparrow$ denotes average percentage improvement than AP in 4 domains. By considering context, $cnn+ml$ and $atn+ml$ achieved better results than AP, which only uses aspect phrase embeddings. After adding nonlinear transformation module, $cnn+mlp+ml$ and $atn+mlp+ml$ further improve the performance. Under the same condition, $atn$ is superior to $cnn$ for our task.

Naturally, aspect phrases in some domains, such as MP3, may have fixed meanings. As a result, an aspect phrase and its context have less correlation under the grouping task in these domains. Therefore, AP achieves a little better result than the other methods except for ADDML. Overall, $atn$ solves the context representation and $mlp+ml$ provides a better metric learning ability for our model.

\textbf{Similarity Threshold} ~Different similarity thresholds $\eta$ results in different negative sample pairs, and have a certain impact on performance of our model. For obtaining a better threshold, we performed experiments on developing data with different similarity value. Figure 3 presents the result on DC dataset. The performance slowly decrease with the growth of threshold, which is in line with intuitively understanding. For obtaining enough negative samples, we chosen 0.3 as the similarity threshold.

\begin{figure}[!htb]
\center
\includegraphics[width=0.6\textwidth]{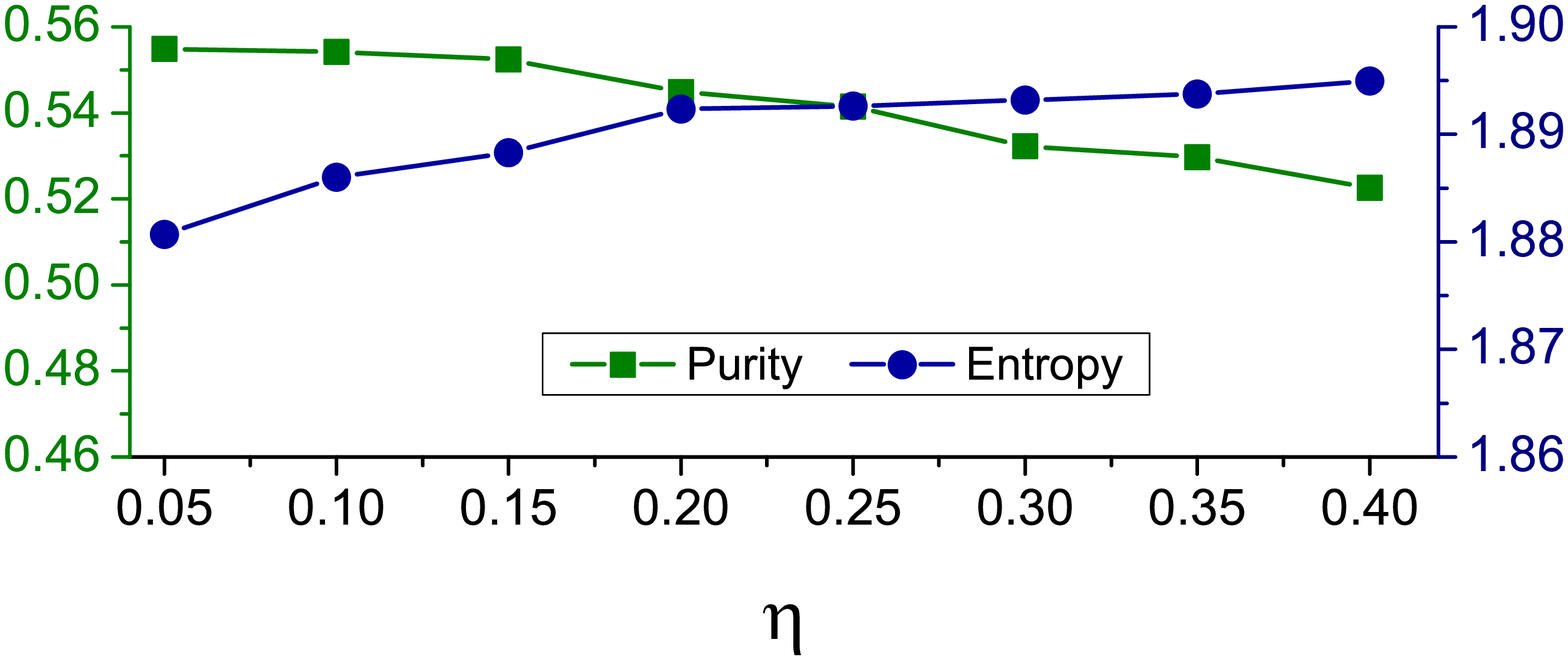}
\caption{Influence of the similarity threshold $\eta$}
\label{fig:example}
\end{figure}

\section{Related Work}
Our work is related to aspect-level sentiment analysis, metric learning, and deep learning. 

For \textbf{aspect-level sentiment analysis}, there are many methods on clustering aspect phrases. Some topic-model-based approaches jointly extract aspect phrases and group them at the same time \cite{ChenMukherjee-1089,MoghaddamEster-1094,LuOtt-1106,JoOh-1388,ZhaoJiang-1111,LinHe-1100}. Those methods tend to discover coarse-grained and grouped aspect phrases, but not specific opinionated aspect phrase themselves. In addition, \newcite{ZhaiLiu-1086} showed that they did not perform well even considering pre-existing knowledge. Some other work focuses on grouping aspect phrases. \newcite{GuoZhu-1250} grouped aspect phrases using multi-level LaSA, which exploits the virtual context documents and semantic structure of aspect phrase. \newcite{ZhaiLiu-1272} used an EM-based semi-supervised learning method for clustering aspect phrases, in which the lexical knowledge is used to provide better initialization for EM. \newcite{ZhaoHuang-1265} proposed a framework of Posterior Regularization to cluster aspect phrases, which formalizes sentiment distribution consistency as a soft constraint. This method requests special semi-structured reviews to estimate the sentiment distribution. In contrast to
these methods, we provide a Siamese neural network to learning feature representation through distance supervising.

\textbf{Metric learning} algorithms have been successfully applied to address the problem of face verification \cite{DingLin-1858,YiLei-1860,CaiWang-1681,HuLu-1683}. A common objective of these methods is to learn a better distance metric so that the distance between a positive pair is smaller than the distance between a negative pair. However, these methods not perform nonlinear transformation. \newcite{HuLu-1683} employed a MLP-based nonlinear transformation, but its input is the given image descriptor, which can be directly concatenated to form feature vectors. In this paper, we adapt this method to the aspect phrase grouping task, and provide an extra attention-based semantic composite model to obtain feature vectors based on word vectors of aspect phrase and its context.  

Our work is related to \textbf{word embedding} and \textbf{deep learning}. Prior research \cite{CollobertWeston-1475,MnihHinton-1476,MikolovSutskever-1335,TangWei-1500,RenZhang-2019} presented different models to improve the performance of word embedding training, and our training is inspried by negative sampling. Deep learning methods \cite{KalchbrennerGrefenstette-1385,Kim-1799,SocherPerelygin-1864,VoZhang-2754,ZhangZhang-2755,ZhangZhang-2756,RenZhang-2021} have been applied to many tasks related to sentiment analysis. In this paper, we explore attention \cite{LuongPham-1792,RushChopra-1794,LingTsvetkov-1793} with a MLP network to tackle the aspect phrase grouping problem. 

\section{Conclusion}
We studied distance metric learning for aspect phrase grouping, exploring a novel deep neural network framework. By leveraging semantic relations between aspect phrase and their contexts, our approach give better performance to strong baselines which achieve the best results in standard benchmark. Our method can be applied to other NLP applications, such as short text clustering and sentence similarity measures.

\section*{Acknowledgement}
This work is supported by the National Natural Science Foundation of China (No. 61173062, 61373108, 61133012), the major program of the National Social Science Foundation of China (No. 11\&ZD189), the key project of Natural Science Foundation of Hubei Province, China (No. 2012FFA088), the Educational Commission of Henan Province, China (No. 17A520050), the High Performance Computing Center of Computer School, Wuhan University and T2MOE201301 from Singapore Ministry of Education.

\bibliographystyle{acl}
\bibliography{addml}

\end{document}